\documentclass[10pt,twocolumn]{article} 
\usepackage{simpleConference}
\usepackage{times}
\usepackage{graphicx}
\usepackage{amssymb}
\usepackage{url,hyperref}

\usepackage{helvet} 
\usepackage{courier}  
\usepackage{graphicx} 
\usepackage{caption} 
\usepackage{bigdelim} 
\usepackage{amssymb}
\usepackage{pifont}
\usepackage{color}
\usepackage{url}
\usepackage{units} 
\usepackage[ruled,noend,vlined]{algorithm2e} 
\usepackage{booktabs}
\usepackage{amsmath}
\usepackage[scientific-notation=true]{siunitx}
\usepackage{skak}
\usepackage{hyperref}
\usepackage[
   style=numeric,
   citestyle=authoryear,
   natbib=true,
   backend=biber,
   maxcitenames=1,
   maxbibnames=99]{biblatex}
\newcommand{\cmark}{\ding{51}}%
\newcommand{\xmark}{\ding{55}}%

\newcommand{\mytitle}{Monte-Carlo Graph Search for AlphaZero}

\hypersetup{
pdftitle={\mytitle},
pdfsubject={Monte-Carlo Graph Search},
pdfauthor={Johannes Czech, Patrick Korus, Kristian Kersting},
pdfkeywords={Monte-Carlo Tree Search,
Graph Search,
Directed Acyclic Graphs,
Epsilon-Greedy Search,
Chess,
Crazyhouse,
Upper Confidence bounds for Trees,
AlphaZero}
}

\begin{document}

\title{\LARGE{\mytitle}}

\author{\textbf{Johannes Czech\textsuperscript{\rm 1},
Patrick Korus\textsuperscript{\rm 1},
Kristian Kersting\textsuperscript{\rm 1, 2, 3}} \\
\\
\textsuperscript{\rm 1}\,Department of Computer Science, Technical University of Darmstadt \\
\textsuperscript{\rm 2}\,Centre for Cognitive Science, Technical University of Darmstadt \\
\textsuperscript{\rm 3}\,hessian.ai — The Hessian Center for Artificial Intelligence \\
Darmstadt, Germany 64289,
\today
\\
\\
\normalsize{
johannes.czech@cs.tu-darmstadt.de, patrick.korus@stud.tu-darmstadt.de, kersting@cs.tu-darmstadt.de} \\
}

\maketitle
\thispagestyle{empty}

\begin{abstract}
\begin{small}
\noindent The \textit{AlphaZero} algorithm has been successfully applied in a range of discrete domains, most notably board games. It utilizes a neural network, that learns a value and policy function to guide the exploration in a Monte-Carlo Tree Search. 
Although many search improvements have been proposed for Monte-Carlo Tree Search in the past, most of them refer to an older variant of the Upper Confidence bounds for Trees algorithm that does not use a policy for planning.
We introduce a new, improved search algorithm for \textit{AlphaZero} which generalizes the search tree to a directed acyclic graph.
This enables information flow across different subtrees and greatly reduces memory consumption.
Along with Monte-Carlo Graph Search, we propose a number of further extensions, such as the inclusion of $\epsilon$-greedy exploration, a revised terminal solver and the integration of domain knowledge as constraints.
In our evaluations, we use the \textit{CrazyAra} engine on chess and crazyhouse as examples to show that these changes bring significant improvements to \textit{AlphaZero}.
    
\end{small}
\end{abstract}

\section{Introduction}

\begin{figure}[t!]
    \centering
    \includegraphics[width=0.9\columnwidth]{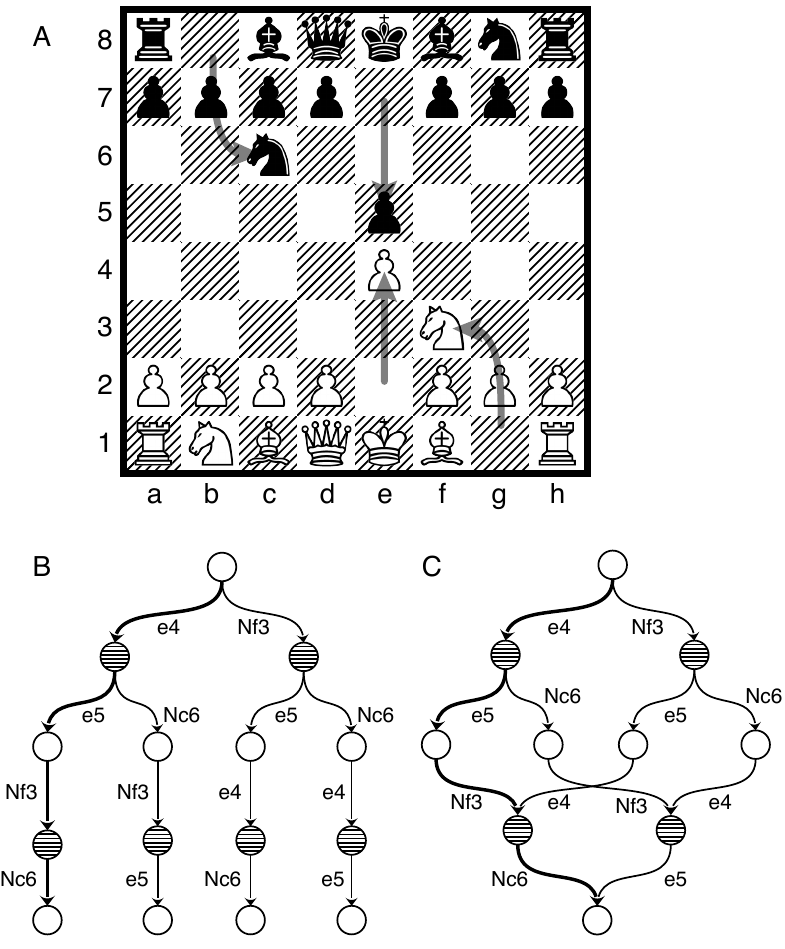}
    \caption{
    It is possible to obtain the King's Knight Opening~(A) with different move sequences~(B, C).
    Trajectories in \textbf{bold} are the most common move order to reach the final position. As one can see, graphs are a much more concise representation.}
    \label{fig:schedules}
\end{figure}

\noindent The planning process of most humans for discrete domains resembles the \textit{AlphaZero} Monte-Carlo Tree Search (MCTS) variant~\citep{silver2017mastering} which uses a guidance policy as its prior and an evaluation function to distinguish between good and bad positions \citep{hassabis2017neuroscience}. The human reasoning process may not be as stoic and structured as a search algorithm running on a computer and they are typically not able to process as many chess board positions in a given time.
Nevertheless, humans are quite good at making valid connections between different subproblems and to reuse intermediate results for other subproblems.
In other words, humans not only look at a problem step by step but are also able to jump between sequence of moves, so called trajectories; they seem to have some kind of \textit{global memory buffer} to store relevant information.
This gives a crucial advantage over a traditional tree search, which does not share information between different trajectories, although an identical state may occur. 

Triggered by this intuition, we generalize the search tree to a Directed Acyclic Graph (DAG), yielding Monte-Carlo Graph Search (MCGS).
The search in our DAG follows the scheme of the Upper Confidence Bound for Trees (UCT) algorithm  \citep{auer2002finite}, but employs a modified forward and backpropagation procedure to cope with the graph structure.
Figure~\ref{fig:schedules} illustrates, how nodes with more than one parent, so called transposition nodes, allow to share information between different subtrees and vastly improve memory efficiency.


More importantly, a graph search reduces the amount of neural network evaluations
that normally take place when the search reaches a leaf node.
Together, MCGS can result in a substantial performance boost, both for a fixed time constraint and for a fixed amount of evaluations.
This is demonstrated in an empirical evaluation on the example of chess and its variant crazyhouse.

Please note, that we intentionally keep our focus here on the planning aspect of \textit{AlphaZero}. We provide a novel and generally applicable planning algorithm that is based on DAGs and comes with a number of additional enhancements. Specifically, our contributions for improving the search of \textit{AlphaZero} are as follows:
\begin{enumerate}
    \item Transforming the search tree into a DAG and providing a backpropagation algorithm which is stable both for low and high simulation counts.
    \item Introducing a terminal solver to make optimal choices in the tree or search graph, in situations where outcomes can be computed exactly.
    \item Combining the UCT algorithm with $\epsilon$-greedy search which helps UCT to escape local optima.
    \item Using Q-value information for move selection
    which helps to switch to the second best candidate move faster if there is a gap in their Q-values.
    \item Adding constraints to narrow the search to events that are important with respect to the domain. In the case of chess these are trajectories that include checks, captures and threats.
\end{enumerate}

Note, that each enhancement, with the exception of the domain knowledge based constraint, can be seen as a standalone improvement and also generalizes to other domains.
We proceed as follows. We start off by discussing related work. Then, we give a quick recap of the PUCT algorithm \citep{rosin2011multi}, the variant of UCT used in \textit{AlphaZero}, and explain our realisation of MCGS and each enhancement individually. 
Before concluding, we touch upon our empirical results. We ran an ablation study showing how each of them increases the performance of \textit{AlphaZero's} planning individually as well as that combining all boosts the performance the most. We explicitly exclude Reinforcement Learning (RL) and use neural network models with pre-trained weights instead. However, we give insight about the stability of the search modifications by performing experiments under different simulation counts and time controls.

\section{Related Work}
There exists quite a lot of prior work on improving UCT search including transposition usage, the Rapid Action Value Estimation (RAVE) heuristic~\citep{gelly2011monte}, move ordering and different parallelization regimes.
A comprehensive overview of these techniques is covered by~\citet{surveryMCTS}.
However, these earlier techniques focus on a version of UCT, which only relies on a value without a policy approximator. Consequently, some of these extensions became obsolete in practice, 
providing an insignificant improvement or
even deteriorated the performance.
Each of the proposed enhancements also increases complexity, and most require a full new evaluation when they are used for PUCT combined with a deep neural network.

\citet{saffidine2012ucd} proposed to use the UCT algorithm with a DAG and suggested an adapted UCT move selection formula~\eqref{eq:node_selection_proposal}. It selects the next move $a_t$ with additional hyperparameters $(d_1, d_2, d_3) \in \mathbb{N}^3$ by
\begin{equation}
    \label{eq:node_selection_proposal}
     a_t = \text{argmax}_a \left(\text{Q}_{d_1}(s_t,a) + U_{d_2, d_3}(s_t, a)\right)\;,
\end{equation}
where
\begin{equation}
     U_{d_2, d_3}(s_t, a) = c_{\text{puct}} \cdot \sqrt{\frac{\log{\sum_b N_{d_2}(s_t,b)}}{N_{d_3}(s_t,a)}}\;.
\end{equation}

The values for values $d_1$, $d_2$ and $d_3$ relate to the respective depth and were chosen either to be 0, 1, 2 or $\infty$. 
Their algorithm was tested on several environments, including a toy environment called \texttt{LEFTRIGHT}, on the board game \texttt{HEX} and on a $6\times6$ \texttt{GO} board using 100, 1000 and 10\,000 simulations respectively.
Their results were mixed. Depending on the game, a different hyperparameter constellation performed best and sometimes even the default UCT-formula, corresponding to $(d_1=0, d_2=0, d_3=0)$, achieved the highest score.


Also the other enhancement suggested in this paper, namely the termaninal solver, extends an already exisiting concept. \citet{chen2018exact} build up on the work by \citet{winands2008monte} and presented a terminal solver for MCTS which is able to deal with drawing nodes and allows to prune losing nodes.

Finally, challenged by the \textit{Montezuma's Revenge} environment, which is a hard exploration problem with sparse rewards, \citet{ecoffet2018montezuma} described an algorithm called \textit{Go-Explore} which remembers promising states and their respective trajectories.
Subsequently, they are able to both return to these states and to robustify and optimize the corresponding trajectory.
Their work only indirectly influenced this paper, but gave motivation to abandon the search tree structure and to keep a memory of trajectories in our proposed methods.


\section{The PUCT Algorithm} 
The essential idea behind the UCT algorithm is to iteratively build a search tree in order to find a good choice within a given time limit \citep{auer2002finite}. Nodes represent states $s_t$ and edges denote actions $a_t$ for each time step $t$. Consider e.\,g.~chess. Here, nodes represent board positions and edges denote legal moves that transition to a new board position. Now, each iteration of the tree search consists of three steps. First, selecting a leaf node, then expanding and evaluating that leaf node and, finally, updating the values and visit counts on all nodes in the trajectory, from the selected leaf node up to the root node. 

To narrow down the search, the PUCT algorithm, introduced by \citet{rosin2011multi} and later refined by \citet{silver2017mastering}, makes use of a prior policy~$\mathbf{p}$. At each state $s_t$, for every time step $t$, a new action $a_t$ is selected according to the UCT-formula~\eqref{eq:node_selection} until either a new unexplored state $s^*$ or a terminal node $s_T$ is reached, i.e., 
\begin{align}
    \label{eq:node_selection}
    a_t = \text{argmax}_a \left(\text{Q}(s_t,a) + U(s_t, a)\right) \; ,\\ \text{where} \quad
    U(s_t,a) = c_{\text{puct}} P(s_t,a) \frac{\sqrt{\sum_b N(s_t,b)}}{1 + N(s_t,a)}\;.
\end{align}
The neural network $f_\theta$ then evaluates the new unexplored state $s^*$.
Every legal action $a_i$ is assigned a policy value $P(s,a_i)$ and the state evaluation $v^*$ is backpropagated along the visited search path.

If we encounter a terminal state $s_T$, then the constant evaluation of either $-1$, $+1$ or $0$ is used instead.
In the case of a two player zero-sum game, the value evaluation $v^*$ is multiplied by $-1$ after each turn. 
The respective Q-values are updated by a Simple Moving Average~(SMA):
\begin{equation}
    \label{eq:q_value_calculation}
    Q'(s_{t},a) = Q(s_{t},a) + \frac{1}{n} \left[ v^* - Q(s_{t},a) \right]\;.
\end{equation}
Unvisited nodes are treated as losses and assigned a value of $-1$.
Moreover, the Q- and U-values are weighted according to the parameter $c_{\text{puct}}$ which is scaled with respect to the number of visits of a particular node:
\begin{equation}
    \label{eq:scale_cpuct}
    c_{\text{puct}}(s) = \log{\frac{\sum_a{N(s,a)}  + c_{\text{puct-base}} + 1}{c_{\text{puct-base}}}} + c_{\text{puct-init}}\;.
\end{equation}
We choose $c_{\text{puct-base}}$ to be
19652 and a $c_{\text{puct-init}}$ value of 2.5 which is based on the choice of \citep{silver2017mastering} but scaled to a larger value range.


\section{Monte-Carlo Graph Search}

\begin{figure}[t]
	\centering
    \includegraphics[width=0.9\columnwidth]{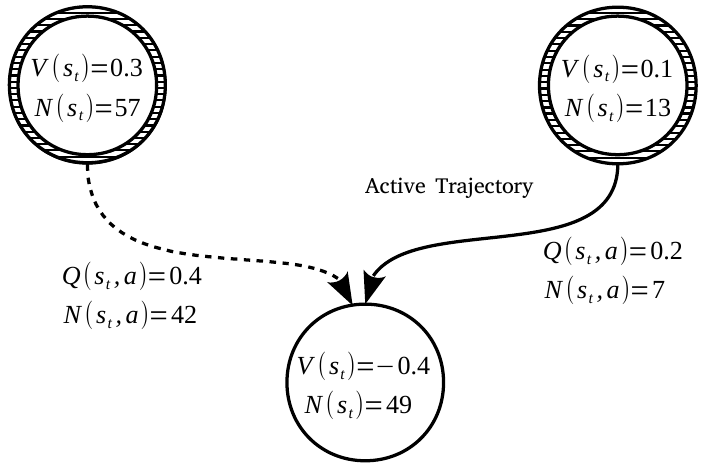}
		\caption{Scenario of accessing a child node on two possible trajectories. The proposed data structure stores the visits and Q-values both on the edges and in the nodes.}
		\label{fig:transposition_showcase}
\end{figure}

The main motivation of having a DAG structure instead of a tree structure is to share computation from different branches and to reduce memory allocation.
If we reach a state, which has already been explored from a different subtree, we can make use of already computed value estimates instead of solving the same subproblem from scratch.

Specifically, we use transpositions and a hash key to detect that the same position was reached on different trajectories. If this is the case, we create an edge between the current node and the pre-existing subtrees.
We incorporate the step counter within the transposition hash key. This way we cannot reach a state with the same key more than once within a trajectory. Therefore, cycles cannot occur.

However, there are other problems to consider. A na\"ive implementation would straightforwardly share information between different subtrees by creating a deep copy of previous neural network evaluation results.
Indeed, this allows reusing previous evaluations and reducing the amount of neural network requests.
Hence, computational resources are not wasted on reevaluating already known information, but instead, new information is gained on each network evaluation. Unfortunately, additional memory is required to copy the policy distribution and value of the preexisting node and the resulting tree structure uses at least as much memory as the vanilla MCTS.
Moreover, the backpropagation can be conducted only on the traversed trajectory or on all possible trajectories on which the transposition is accessible.

Updating all trajectories has both a bad scaling behaviour as well as leading to negative side-effects \citep{saffidine2012ucd}. If we only update the traversed trajectory, we encounter the problem of leaking information instead:
each Q-value of the parent nodes of a tranposition node bases its approximation only on a subset of the value information within the subtree, but uses the full subtree for node selection.
Therefore, it loses the information which is incorporated in the Q-value of the other parent nodes of the transposition node. This phenomenon occurs more frequently as the number of simulations increases and makes this approach unstable.

In the following, we develop a solution for both of the key issues: we address the memory problem, while preventing information leaks. 
To this end, MCGS will be explained and evaluated using PUCT for move selection as an example but could in principle also be applied for UCT or plain MCTS search.

\begin{algorithm}[t]
\SetAlgoLined
\KwData{\texttt{rootNode}, $s_0$, $Q_{\epsilon}$}
\KwResult{\texttt{trajectory}, \texttt{value}}
\texttt{node} $\leftarrow$  \texttt{rootNode}\;
$s_t \leftarrow s_0$\;
 \While{\texttt{node} is not \textit{leaf}}{
 (\texttt{nextNode}, \texttt{edge}) $\leftarrow$ select node using~\eqref{eq:node_selection}\;
 append (\texttt{node}, \texttt{edge}) to \texttt{trajectory}\;
 \If{\texttt{nextNode} is \textit{transposition}}{
 $Q_{\delta} \leftarrow  Q(s_t, a) - V^*(s_{t+1})$\;
 \If{$Q_{\delta} > Q_{\epsilon}$}{
 $Q_\phi(s_t, a) \leftarrow N(s_t, a) \cdot Q_\delta(s_t, a) + V^*(s_{t+1})$\;
 $Q'_\phi(s_t, a) \leftarrow \max({V_\text{min}},\min(Q_\phi(s_t, a),V_\text{max}))$\;
 \texttt{value} $\leftarrow Q'_{\phi}$\;
 \textbf{return} \texttt{trajectory}, \texttt{value}\;
 }
 }
 \If{\texttt{nextNode} is \textit{terminal}}{
    \texttt{value} $\leftarrow$ \texttt{nextNode}.value\;
    \textbf{return} \texttt{trajectory}, \texttt{value}\;
 }
 
  \texttt{node} $\leftarrow$  \texttt{nextNode}\;
  $s_t \leftarrow$ apply action \texttt{edge.}$a$ on $s_t$\;
}
expand \texttt{node}\;
$(\text{\texttt{node.}}v, \text{\texttt{node.}}\mathbf{p}) \leftarrow f_\theta(s_t) $\;
\texttt{value} $\leftarrow \texttt{node.}v$\;
\textbf{return} \texttt{trajectory}, \texttt{value}\;
\caption{Node Selection and Expansion of MCGS}
\label{code:fwd_mcgs}
\end{algorithm}

\begin{algorithm}[t]
\SetAlgoLined
\KwData{\texttt{trajectory}, \texttt{value}}
\KwResult{updated search graph}
\texttt{qTarget} $\leftarrow \infty$\;
\While{(\texttt{node}, \texttt{edge}) in \textbf{reverse}(\texttt{trajectory})}{
\eIf{\texttt{qTarget} != $\infty$}{
 $Q_{\delta} \leftarrow  Q(s_t, a) - $ \texttt{qTarget}\;
 $Q_\phi(s_t, a) \leftarrow N(s_t, a) \cdot Q_\delta(s_t, a) + V^*(s_{t+1})$\;
 $Q'_\phi(s_t, a) \leftarrow \max({V_\text{min}},\min(Q_\phi(s_t, a),V_\text{max}))$\;
 \texttt{value} $\leftarrow Q'_{\phi}$\;
}
{
\texttt{value} $\leftarrow$ $-$\texttt{value}\;
}
update \texttt{edge.}$q$ with \texttt{value}\;
\texttt{edge.}$n$\texttt{+}\texttt{+}\;
update \texttt{node.}$v$ with \texttt{value}\;
\texttt{node.}$n$\texttt{+}\texttt{+}\;
\eIf{\texttt{node} is transposition}{
\texttt{qTarget} $\leftarrow$ $-$\texttt{node.}$v$\;
}
{
\texttt{qTarget} $\leftarrow \infty$
}
}

\caption{Backpropagation of MCGS}
\label{code:backup_mcgs}
\end{algorithm}

\subsection{Data-Structure} 
There are multiple possible data structures to realize a DAG and to conduct backpropagation in a DAG. Q-values and visits can be stored on the edges, on the nodes or even both.

We propose to store the Q-values on both to avoid information leaking.
Indeed, this is accompanied with a higher memory consumption and a higher computational effort.
However, it also allows us to differentiate between the current believe of the Q-value on the edge and the more precise value estimation on the transposition node.
The value estimation of the next node is generally more precise or has the same precision than the incoming Q-value on the edge because $N(s_t, a) \leq N(s_{t+1})$.

As our data structure, we keep all trajectories of the current mini-batch in memory, because these will later be used during the backpropagation process.

\subsection{Selection, Expansion and Backpropagation}
With the data structure at hand, we can now realise the value update as well as modify backpropagation correspondingly. Consider the situation depicted in Figure~\ref{fig:transposition_showcase}.
If the next node of a simulation trajectory is a transposition node, i.e., a node that has more than one parent node, we define our target $V^*(s_t+1)$ for the Q-value $Q(s_t, a)$ as the inverse of the next value\footnote{In case of a single player environment, $V^*(s_t+1)$ is equivalent to $V(s_t+1)$.} estimation:
\begin{equation}
    V^*(s_t) = {-V(s_{t})}.
\end{equation}
Now, we calculate the residual of our current Q-value belief $Q(s_t, a)$ compared to the more precise value estimation $V^*(s_{t+1})$ thereafter.
We define this residual as $Q_{\delta}$:
\begin{equation}
\label{eq:q_delta}
    Q_{\delta}(s_t, a) = Q(s_t, a) - V^*(s_{t+1}),
\end{equation}
that measures the amount of the current information leak.

If our target value $V^*(s_{t+1})$ has diverged from our current belief $Q(s_t, a)$, e.\,g. $|Q_{\delta}| > 0.01$, we already have a sufficient information signal and do not require an additional neural network evaluation.
Consequently, we stop following the current trajectory further and avoid expensive neural network evaluations which are unlikely to provide any significant information gain.
We call $Q_{\epsilon}$ the hyper-parameter for the value of $0.01$ and it remains the only hyper parameter in this algorithm.

If, however,  $|Q_{\delta}| \leq Q_{\epsilon}$, then we iteratively apply the node selection formula \eqref{eq:node_selection} of the PUCT algorithm to reach a leaf-node.
Otherwise, in case of $|Q_{\delta}| > Q_{\epsilon}$, we backpropagate a value that does not make use of a neural network evaluation and brings us close to $V^*(s_{t+1})$.
This correction value, denoted as $Q_\phi(s_t, a)$, is 
\begin{equation}
    \label{eq:q_phi_1}
    Q_\phi(s_t, a) = N(s_t, a) \cdot Q_\delta(s_t, a) + V^*(s_{t+1})
\end{equation}
and can become greater than $Q_\text{max}$ or smaller than $Q_\text{min}$ for large $N(s_t, a)$.
To ensure that we backpropagate a well-defined value, we clip our correction value to be within $[V_{min}, V_{max}]$, i.e., 
\begin{equation}
    \label{eq:q_phi_2}
    Q'_\phi(s_t, a) = \max({V_\text{min}},\min(Q_\phi(s_t, a),V_\text{max})).
\end{equation}
This clipping process is similar to the gradient clipping \citep{Zhang2020Why} procedure, which is often used in Stochastic Gradient Descent (SGD). We also just 
incorporate the correction procedure of \eqref{eq:q_phi_1} and \eqref{eq:q_phi_2} into the backpropagation process after every transposition node. 
A compact summary of the forward- and backpropagation procedure is shown in Algorithm \ref{code:fwd_mcgs} and \ref{code:backup_mcgs}.

\subsection{Discussion}
The MCGS algorithm as described here makes several assumptions. First, we assume the state to be Markovian, because the value estimation of trajectories with an alternative move ordering is shared along transposition nodes.
This assumption, however, might be violated if a transposition table is used to store neural network evaluations. In practice, for environments such as chess that are theoretically Markovian but can make use of history information in the neural network input representation, this did not result in a major issue.

As previously mentioned, our data-structure employs redundant memory and computation for nodes that are currently not transposition nodes but may become transposition nodes in the future. However, it should be considered that the bottlekneck of \textit{AlphaZero's} PUCT algorithm is actually the neural network evaluation, typically executed on a Graphics Processing Unit (GPU). This may explain our observation that spending a small overhead on more CPU computation did not result in an apparent speed loss.

Bare in mind, that a node allocates memory for several statistics. While value and visits are scalar, a node also has to hold the policy distribution for every legal move even if it will never be used through the search. The memory consumption of the latter is larger by orders of magnitude. As a consequence, we observe a memory reduction of 30\,\% to 70\,\% depending on the position and the resulting amount of transposition nodes. 
















\section{Further Enhancements of MCGS}
Beyond the MCGS algorithm, we propose a set of additional independent enhancements to the \textit{AlphaZero} planning algorithm. Each of them brings an improvement to both the original tree search as well as MCGS. In the following, we describe these methods in detail and in the evalutation, we investigate how a combination of all proposed methods performs compared to each enhancement, individually.

\subsection{Improved Discovering of Forcing Trajectories}

\begin{table*}[t]
\fontsize{9}{11}
\centering
\caption{Comparison of different terminal solver implementations.}
\label{tab:terminal_solver}
\begin{tabular}{ c c c c}
\toprule
  & \textbf{Exact-win-MCTS 2.0 (Ours)} & \textbf{Exact-win-MCTS} & \textbf{MCTS-Solver+MCTS-MB} \\
\midrule
 Node States & \texttt{WIN}, \texttt{LOSS}, \texttt{DRAW}, \texttt{UNKNOWN}
 & \texttt{WIN}, \texttt{LOSS}, \texttt{DRAW}, \texttt{UNKNOWN} & \texttt{WIN} and \texttt{LOSS} \\
 Optional Node States &  \texttt{TB\_WIN}, \texttt{TB\_LOSS}, \texttt{TB\_DRAW} & - & - \\
 Member Variables & \texttt{UNKNOWN\_CHILDREN\_COUNT}, & \texttt{UNKNOWN\_CHILDREN\_COUNT} & - \\
  & \texttt{END\_IN\_PLY} & & \\
 Nodes have been proven & Prune & Prune & May revisit \\
 Search in vain & No & No & Maybe \\
 Draw games & \cmark & \cmark & \xmark \\
 Supports Tablebases & \cmark & \xmark & \xmark \\
 Selects shortest mate & \cmark & \xmark & \xmark \\
 Simulation results & Remain unchanged & Remain unchanged & May be changed \\
\bottomrule
\end{tabular}
\end{table*}
\normalsize


Due to sampling employed, MCGS may miss a critical move or sample a move again and again even if it has been explored already with an exact ``game-theoretical'' value. 
To help pruning losing lines completely and increasing the chance of finding an exact winning move sequence, we now introduce a terminal solver into the planning algorithm \citep{chen2018exact}. 
Doing so allows early stopping, and to select the so far known shortest line when in a winning position. It also provides a stronger learning signal during RL.
Respectively, the longest line can be chosen, when in a losing position, and if a step counter is computed.
\begin{algorithm}[b!]
\SetAlgoLined
 \If{has-loss-child}{
 mark \texttt{WIN}\;
  parent.\texttt{UNKNOWN\_CHILDREN\_COUNT-}\texttt{-}\;
  \texttt{END\_IN\_PLY} $\leftarrow  \min_{\text{child}}$($\texttt{END\_IN\_PLY}_{\text{child}}$) + 1\;
  }
  \If{\texttt{UNKNOWN\_CHILDREN\_COUNT} == 0}{
  \eIf{has-draw-child}{
     mark \texttt{DRAW}\;
     parent.\texttt{UNKNOWN\_CHILDREN\_COUNT-}\texttt{-}\;
     \texttt{END\_IN\_PLY} $\leftarrow \min_{\text{child}}$($\texttt{END\_IN\_PLY}_{\text{child}}$) + 1\;
  }
  {
 mark \texttt{LOSS}\;
 parent.\texttt{UNKNOWN\_CHILDREN\_COUNT-}\texttt{-}\;
 \texttt{END\_IN\_PLY} $ \leftarrow \max_{\text{child}}$($\texttt{END\_IN\_PLY}_{\text{child}}$) + 1;
  }
}
\caption{Backpropagation of Exact-win-MCTS~2.0}
\label{code:backprop_terminal_solver}
\end{algorithm}

Specifically, we add an identifier called \texttt{END\_IN\_PLY}, which keeps track of the number of steps until a terminal node is reached.
When a new node has been solved, \texttt{END\_IN\_PLY} is assigned the same value as a designated child node and is then incremented by one.
In case of a \texttt{LOSS}, the terminal solver chooses the child node with the highest \texttt{END\_IN\_PLY} value and the smallest value in case of a \texttt{WIN} or \texttt{DRAW}.
Besides that, we add three new node states \texttt{TB\_WIN}, \texttt{TB\_LOSS}, and \texttt{TB\_DRAW}, which are used to determine forced lines to reach a table base position.
If a node has been proven to be a \texttt{TB\_LOSS}/\texttt{LOSS}, then we prune its access by setting its Q-value to $-\infty$ and policy value to $0.0$.

As argued in Table~\ref{tab:terminal_solver}, this terminal solver gives a significant advantage over other terminal solvers. 
The pseudo-code of the backpropagation for determining the node sates \texttt{WIN}, \texttt{LOSS} and \texttt{DRAW} is shown in Algorithm \ref{code:backprop_terminal_solver}.
The computation of solving node states that can reach a table base position by force is carried out analogously to the aforementioned one.
As soon as we reach a tablebase position, we can still make use of our value evaluation of our neural network model, in order to converge to a winning terminal node faster.
If we only used the tablebase value evaluation instead, we would encounter the problem of reaching a plateau and our search would be unable to differentiate between different winning or losing node states.

In order to accelerate the convergence to terminal nodes, we decouple the forward and backpropagation process of terminal trajectories and allow a larger amount of terminal trajectories during the creation of a mini-batch.











\subsection{Random Exploration to Avoid Local Optima}
The $\epsilon$-greedy search is a well known exploration mechanism which is often applied in RL algorithms such as Q-learning~\citep{watkins1992q}.
The idea behind \mbox{$\epsilon$-greedy} search is to follow the so far best known trajectory and to explore a different action with a probability $\epsilon_\text{greedy}$ instead.
Over time, the influence of the prior policy in the PUCT algorithm diminishes and more simulations are spend on the action with the current maximum Q-value. Formula \eqref{eq:node_selection} is meant to provide a mechanism to balance exploiting known values and exploring new, possibly unpromising nodes. In fact, it is proven that this formula will find an optimal solution with a sufficient amount of simulations \citep{auer2002finite}.

However, in practice, we see that a lot of the times the fully deterministic PUCT algorithm from \textit{AlphaZero} needs an unpractical amount of simulations to revisit optimal actions where the value evaluations of the first visits are misleading. This motivates adding stochasticity to sooner escape such scenarios. Breaking the rule of full determinism and employing some noise is also motivated by optimization algorithms such as SGD \citep{srivastava2014dropout} as well as dropout~\citep{srivastava2014dropout} that improve convergence and robustness. In the context of RL it was suggested by \citet{silver2017mastering} to apply Dirichlet noise $\text{Dir}(\alpha)$ on the policy distribution $\mathbf{p}$ of the root node to increase exploration. However, this technique has several disadvantages when being employed in a tournament setting particularly with large simulation budgets.

Utilizing a static Dirichlet noise at the root node generally increases the exploration rate of a few unpromising actions because the amount of suboptimal actions is often larger than relevant actions. Therefore, it is more desirable to apply uniform noise. Additionally, it is favorable to apply such noise not only at root level but at every level of the tree. Such additional uniform noise is what underlies the $\epsilon$-greedy algorithm.

PUCT, UCT and $\epsilon$-greedy have a lot in common by trying to find a good compromise for the
exploration-exploitation dilemma. 
PUCT and UCT usually converge faster to the optimal trajectory than $\epsilon$-greedy~\citep{sutton2018reinforcement}. $\epsilon$-greedy selects actions greedily but with static uniform noise, instead. Therefore, it provides a suitable mechanism to overcome local optima where PUCT gets stuck.
The $\epsilon$-greedy algorithm can be straightforwardly implemented at the root level of the search as there is no side-effect of sampling a random node from the root node, except of potentially wasting simulations on unpromising actions. However, if the mechanism is utilized at nodes deeper in the tree, we disregard the value expectation formalism and corrupt all its parent nodes on its trajectory.

Following this regime, we propose to use disconnected trajectories for $\epsilon$-greedy exploration.
In other words, we intentionally create an information leak.

Specifically, a new exploration trajectory is started if a random variable uniformly drawn from $[0, 1]$ is $\leq \epsilon$.
Next, we determine the depth on which we want to start the branching.
We want to generally, prefer branching at early layers.
Therefore, we draw a new random variable $r_2$ and exponentially reduce the chance of choosing a layer with increasing depth
\begin{equation}
    \label{eq:depth_selection}
    \text{depth} = - \log_2(1 - r_2) - 1\;.
\end{equation}


Unexplored nodes are expanded in descending order.
Usually, the policy is ordered already, to allow a more efficient dynamic memory allocation and node-selection formula.
Therefore this step does not require an additional overhead.





\subsection{Using Q-Value information for Move Selection}

The default move selection policy $\pi$ is based on the visits distribution of the root node which can optionally be adjusted by a temperature parameter~$\tau$,

\begin{equation}
    \pi ( a | s_0 ) = {N(s_0 , a )^{\frac{1}{\tau}}}\slash{\Big(\sum\nolimits_b N(s_0 , b)^{\frac{1}{\tau}}\Big)}.
\end{equation}

Including the Q-values for move selection can be beneficial because the visits and Q-value distribution have a different convergence behaviour.
The visit distribution increases linearly over time whereas the Q-value can converge much faster if the previously best move found was found to be losing.
In \citep{crazyara_frontiers}, it was proposed to use a linearly combination of the visits and Q-values to build the move selection policy~$\pi$.
Now, we choose a more conservative approach and only inspect the action with the highest and second highest visits count
which we label $a_\alpha$ and $a_\beta$ respectively.
Next, we calculate the difference in their Q-values
\begin{equation}
    Q_\Delta(s_0,a_\alpha, a_\beta ) = Q(s_0, a_\beta) - Q(s_0, a_\alpha),
\end{equation}

and if $Q_\Delta(s_0,a_\alpha, a_\beta )$ was found to be $>$\.0, then we boost $\pi( a_\beta, | s_0) $ by $\pi_\text{corr}(s_0, a_\alpha, a_\beta)$
\begin{equation}
    \pi( a_\beta, | s_0)' = \pi( a_\beta, | s_0) + \pi_\text{corr}(s_0, a_\alpha, a_\beta)\;,
\end{equation}
where
\begin{equation}
    \pi_\text{corr}(s_0, a_\alpha, a_\beta) = Q_\text{weight} Q_\Delta(s_0,a_\alpha, a_\beta ) \pi( a_\alpha, | s_0),
\end{equation}
and re-normalize $\pi$ afterwards.
$Q_\text{weight}$ acts as an optional weighting parameter which we set to $2.0$ because our Q-values are defined to be in $[-1,+1]$.

This technique only involves a small constant overhead and helps to switch to the second candidate move faster, both in tournament conditions as well as when used in the target distribution during RL.



\subsection{Incorporating Domain Knowledge through Constraints} 

It is a common advice in chess --- both on amateur and master level --- to first explore moves that are checks, captures and threats.
Hence, we add the constraint to first explore all checking moves before other moves during $\epsilon$-greedy exploration.
We again choose a depth according to \eqref{eq:depth_selection} and follow the so far best known move sequence for expanding a so far unvisited checking move.

The checking moves are being explored according to the ordering of the neural network policy, which makes the exploration of more promising checking moves happen earlier.
After all checking moves have been explored, the node is assigned a flag and the remaining unvisited nodes are expanded.
After all moves have been expanded, the same procedure as for $\epsilon$-greedy search is followed.
As above, we disable backpropagation for all preceding nodes on the trajectory when expanding a check node and choose an $\epsilon_{check}$ value of $0.01$.
In this scenario, we test the expansion of checking moves but it could be extended to other special move types, such as capture or threats as well.

A benefit of this technique is that it provides the guarantee, that all checking at earlier depth are explored quickly, even when the neural network policy fails to acknowledge them, and without influencing the value estimation of its parent nodes.






\section{Empirical Evaluation}

\begin{figure}[t]
    \centering
    \includegraphics[width=0.9\columnwidth]{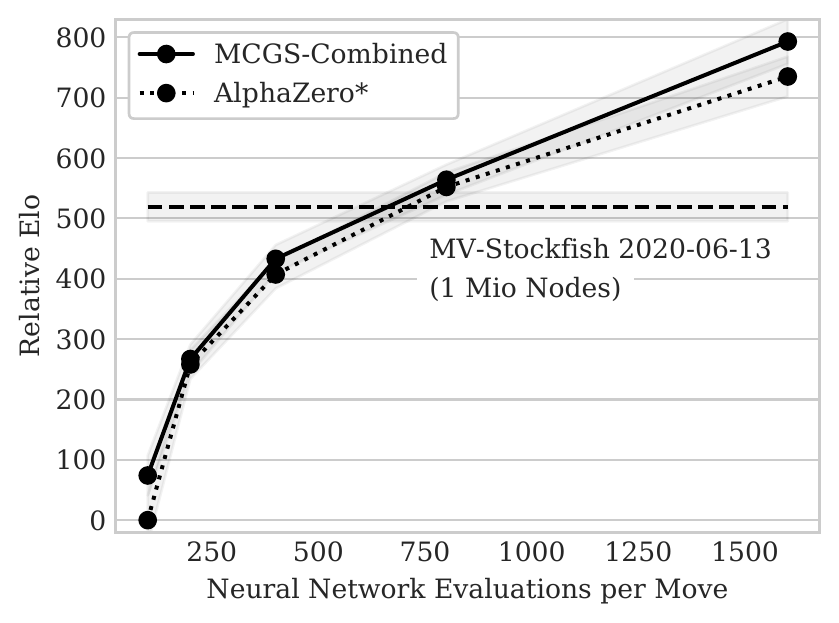}%
    \caption{Elo development relative to the number of neural network evaluations in \textbf{crazyhouse}.}
    \label{fig:elo_increase_cz}
\end{figure}

Our intention here is to evaluate the benefits of our MCGS and the aforementioned search modifications empirically.
In particular, we want to investigate whether each of the contributions boost the performance of \textit{AlphaZero's} planning individually and whether a combination is beneficial.

In our evaluation we use pre-trained convolutional neural network models.
For crazyhouse we use a model of the \texttt{RISEv2}~\citep{crazyara_frontiers} architecture which was first trained on $\approx\,$0.5 million
human games of the lichess.org database and subsequently improved by 436$\pm$30 Elo over the course of $\approx\,$2.37 million self-play games
\citep{czech2019deep}.

The same network architecture was then employed for chess and 
trained on the free Kingbase~2019 dataset \citep{Havard2019} with the same training setup as in \citep{crazyara_frontiers}.
After convergence the model scored a 
move validation accuracy of 57.2\,\% and a validation mean-squared-error of 0.44 for the value loss.
\begin{figure}[t]
    \centering
    \includegraphics[width=0.9\columnwidth]{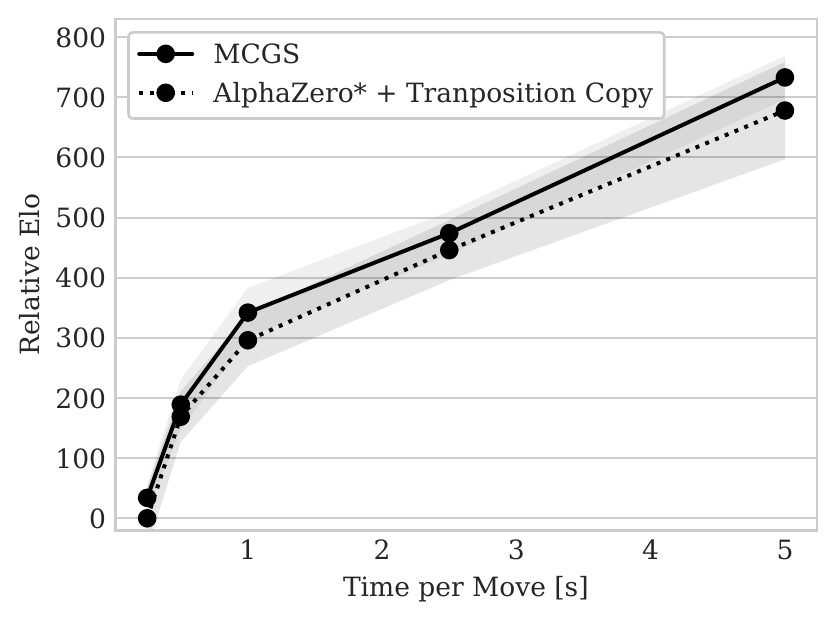}
    \caption{Elo development  in \textbf{crazyhouse} over time of MCGS compared to MCTS which uses a hash table as a transposition buffer to copy neural network evaluations.}
    \label{fig:mcgs_eval}
\end{figure}
This re-implementation is labeled as \textit{AlphaZero*} because of not being the original implementation and using different set of hyperparameters values\footnote{All technical details, such as hyperparameter values and hardware are given in the Appendix, Section ~\ref{sec:appendix}.}

One of the changes compared to \textit{AlphaZero} is the avoidance of Dirchlet noise by setting $\epsilon_\text{Dir}$ to \num{0.0} and the usage of $\text{Node}_\tau$ which flattens the policy distribution of each node in the search tree.
The hardware configuration, used in our experiments, achieves about 17\,000 neural network evaluations per second for the given neural network architecture.

In our first experiment as as shown in Figure~\ref{fig:mcgs_eval}, we compare the scaling behaviour in crazyhouse between our presented MCGS algorithm and a re-implementation of the \textit{AlphaZero} algorithm that also makes use of a transposition table to store neural network evaluations.
We observe that MCGS outperformed the transposition look-up table approach across all time controls, demonstrating that MCGS can be implemented efficiency and excels by providing a more expressive information flow along transposition nodes.
In particular, it becomes apparent, that the Elo gap between the two algorithms slightly increases over time, suggesting an even better performance in the long run or when executed on stronger hardware.

\begin{figure}[t]
    \centering
    \includegraphics[width=0.9\columnwidth]{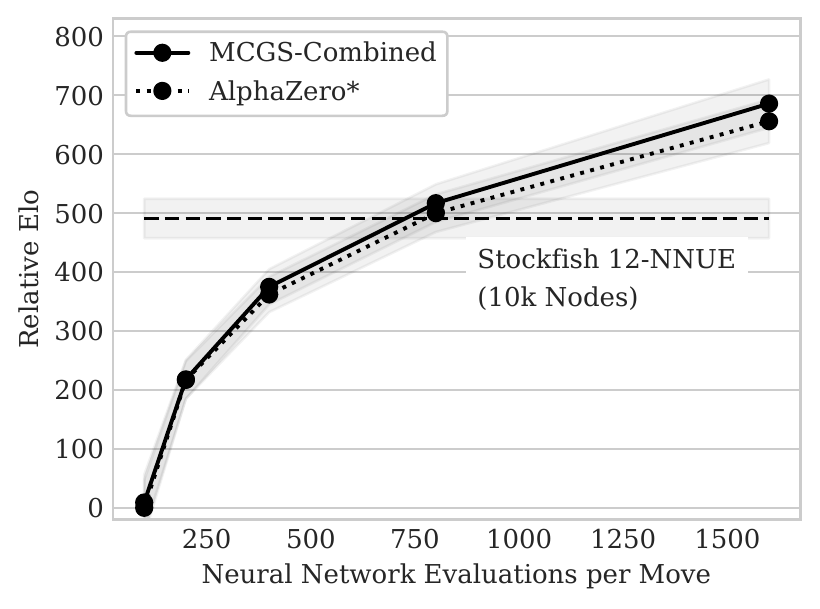}
    \caption{Elo development relative to the number of neural network evaluations in \textbf{chess}.}
    \label{fig:elo_increase_chess}
\end{figure}

Next, we investigate the scaling behaviour relative to the number of neural network evaluations, both for crazyhouse and chess in Figure \ref{fig:elo_increase_cz} and \ref{fig:elo_increase_chess}.
Again, we can draw a similar conclusion.
For a small amount of nodes, the benefit of using all MCGS with all enhancements over \textit{AlphaZero*} is relatively small, but increases the longer the search is performed.
We state the number of neural network evaluations per move instead of the number of simulations because this provides a better perspective on the actual run-time cost.
Terminal visits as well as the backpropagation of correction values $Q_\phi(s_t, a)$  as shown in \eqref{eq:q_phi_2} can be fully executed on CPU during the GPU computation.

To put the results into perspective we also add the performance of \textit{Multi-Variant Stockfish}\footnote{{github.com/ddugovic/Stockfish}} \textit{2020-06-13}~\ref{fig:elo_increase_cz} in Figure using one million nodes per move.
For chess we use the official \textit{Stockfish 12} release with 10\,000 nodes per move as our baseline, which uses a NNUE-network~\citep{yu_ue_2018}\footnote{nn-82215d0fd0df.nnue} as its main evaluation function.
The playing strength between the MCGS for crazyhouse and chess greatly differs in strength, however, this is primarily attributed to the model weights and not the search algorithm itself.

\begin{figure}[t]
    \centering
    \includegraphics[width=0.9\columnwidth]{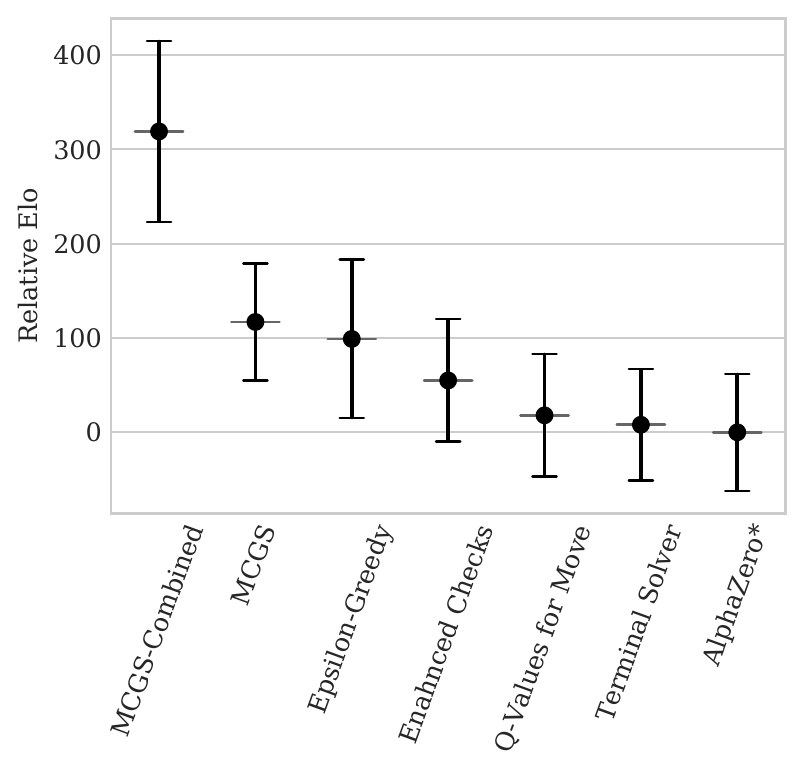}
    \caption{Elo comparison of the proposed search modification in \textbf{crazyhouse} using \textbf{five seconds per move}. On the used hardware this resulted in 100\,000 - 800\,000 total nodes per move.}
    \label{fig:elo_5s_crazyhouse}
\end{figure}

\begin{figure}[t]
    \centering
    \includegraphics[width=0.9\columnwidth]{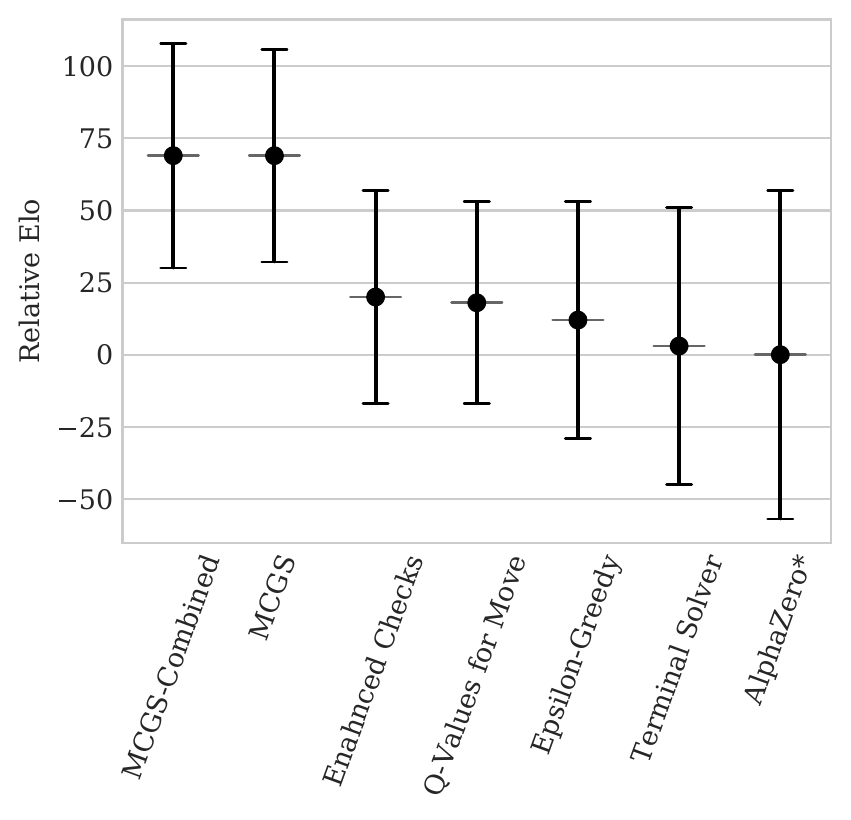}
    \caption{Elo comparison of the proposed search modification in \textbf{chess} using \textbf{five seconds per move}.}
    \label{fig:elo_5s_chess}
\end{figure}
For evaluating the performance in chess, we use a list of 2514 unique starting positions\footnote{{\scriptsize{sites.google.com/site/gaviotachessengine/download/gaviota-starters.pgn.zip}}
}
to have a more diverse range of positions and less amount of draws.
Additionaly, we enable the usage of 3-4-5 Syzygy tablebases.
In crazyhouse we face the opposite problem. The first player is given a much higher advantage from the initial starting position and the chance of drawing games is significantly smaller.
Therefore, we gathered a list of 691 starting positions\footnote{
\scriptsize{github.com/ianfab/books/blob/master/crazyhouse\_mix\_cp\_130.epd}
} which both the \textit{CrazyAra 0.7.0} and  \textit{Multi-Variant Stockfish~11} engine evaluated to be within [-130, +130] centi-pawns.
At last we evaluate each search modification individually both for crazyhouse, as seen in Figure~\ref{fig:elo_5s_crazyhouse} and chess, as shown in Figure~\ref{fig:elo_5s_chess}.
For the example of crazyhouse, it builds a tree of 100\,000 - 800\,000 nodes per move in this given time control.
The amount of nodes per game generally increase over the course of a game because the subtree of the previous search is reused for proceeding searches. 
Each individual search modification appears to improve the performance, whereas using MCGS instead of a tree structure yields the greatest benefit.
In crazyhouse, MCGS resulted in $\approx$\,+110 Elo followed by $\epsilon$-greedy with $\approx$\,+100 Elo.
Combining all enhancement at once, which we refer to as {MCGS-Combined}, leads to $\approx$\,+310 Elo and greatly surpassed each variant individually.
In the case of chess, the impact is not as evident as for crazyhouse but we also recognize an improvement
of $\approx$\,+69 Elo when using 
all search enhancements at once.




















\section{Conclusions}
Our experimental results clearly show that using DAGs instead of trees significantly increases the efficiency and performance of the search in \textit{AlpaZero}. Each individual enhancement that we propose gives better results, but the combination exceeds them all and remains stable even if the graph gets very large in long planning procedures. Together, they boost the performance of \textit{CrazyAra}, which is the current state-of-the-art in crazyhouse.

For classic chess we see less drastic, but still respectable improvements given the models, that were learned through supervised learning on human expert games.
Additionally, our results suggest that MCGS gains in value, the longer the search is executed or on stronger hardware.
Beyond that, the proposed techniques are generally applicable and also work outside the \textit{AlphaZero} framework.

Our work provides several interesting avenues for future work. MCGS should be further improved by sharing key trajectories between nodes that are not an exact transposition but similar to each other.
Furthermore, one should move beyond the planning setting and explore, how MCGS effects the learning through self-play in an RL setting and how it will impact the performance in chess and other environments.

\newpage

\section{Acknowledgements}
The authors thank Matuiss\footnote{{github.com/Matuiss2}} for continuously testing the \textit{CrazyAra} engine and giving constructive feedback. The move generation, terminal and Syzygy-tablebase validation procedures in \textit{CrazyAra-0.8.4} of the crazyhouse and chess environment have been integrated from \textit{Multi-Variant-Stockfish}. We thank Daniel Dugovic, Fabian Fichter, Niklas Fiekas and the entire Stockfish developer community for providing a fast and stable open-source implementation of these algorithms. We also thank Moritz Willig and our other reviewers for proofreading this paper. Finally, we acknowledge the support of the EU project TAILOR (grand agreement 952215) under ICT-48-2020.







\printbibliography

@article{saffidine2012ucd,
  title={UCD: Upper Confidence bound for rooted Directed acyclic graphs},
  author={Saffidine, Abdallah and Cazenave, Tristan and M{\'e}hat, Jean},
  journal={Knowledge-Based Systems},
  volume={34},
  pages={26--33},
  year={2012},
  publisher={Elsevier}
}

@ARTICLE{surveryMCTS,
  author={C. B. {Browne} and E. {Powley} and D. {Whitehouse} and S. M. {Lucas} and P. I. {Cowling} and P. {Rohlfshagen} and S. {Tavener} and D. {Perez} and S. {Samothrakis} and S. {Colton}},
  journal={IEEE Transactions on Computational Intelligence and AI in Games}, 
  title={A Survey of Monte Carlo Tree Search Methods}, 
  year={2012},
  volume={4},
  number={1},
  pages={1-43},
  doi={10.1109/TCIAIG.2012.2186810}
}

@article{ecoffet2018montezuma,
  title={Montezuma's revenge solved by go-explore, a new algorithm for hard-exploration problems (sets records on pitfall, too)},
  author={Ecoffet, Adrien and Huizinga, Joost and Lehman, Joel and Stanley, Kenneth O and Clune, Jeff},
  journal={Uber Engineering Blog, Nov},
  year={2018}
}

@ARTICLE{crazyara_frontiers,
AUTHOR={Czech, Johannes and Willig, Moritz and Beyer, Alena and Kersting, Kristian and F\"urnkranz, Johannes},  
TITLE={Learning to Play the Chess Variant Crazyhouse Above World Champion Level With Deep Neural Networks and Human Data},      
JOURNAL={Frontiers in Artificial Intelligence},      
VOLUME={3},      
PAGES={24},     
YEAR={2020},      
URL={https://www.frontiersin.org/article/10.3389/frai.2020.00024},       
DOI={10.3389/frai.2020.00024},      
ISSN={2624-8212}
}

@article{silver2017mastering,
  title={Mastering the game of go without human knowledge},
  author={Silver, David and Schrittwieser, Julian and Simonyan, Karen and Antonoglou, Ioannis and Huang, Aja and Guez, Arthur and Hubert, Thomas and Baker, Lucas and Lai, Matthew and Bolton, Adrian and others},
  journal={nature},
  volume={550},
  number={7676},
  pages={354--359},
  year={2017},
  publisher={Nature Publishing Group}
}

@inproceedings{chen2018exact,
  title={Exact-win strategy for overcoming AlphaZero},
  author={Chen, Yen-Chi and Chen, Chih-Hung and Lin, Shun-Shii},
  booktitle={Proceedings of the 2018 International Conference on Computational Intelligence and Intelligent Systems},
  pages={26--31},
  year={2018}
}

@inproceedings{winands2008monte,
  title={Monte-Carlo tree search solver},
  author={Winands, Mark HM and Bj{\"o}rnsson, Yngvi and Saito, Jahn-Takeshi},
  booktitle={International Conference on Computers and Games},
  pages={25--36},
  year={2008},
  organization={Springer}
}

@book{sutton2018reinforcement,
  title={Reinforcement learning: An introduction},
  author={Sutton, Richard S and Barto, Andrew G},
  year={2018},
  publisher={MIT press}
}

@article{hassabis2017neuroscience,
  title={Neuroscience-inspired artificial intelligence},
  author={Hassabis, Demis and Kumaran, Dharshan and Summerfield, Christopher and Botvinick, Matthew},
  journal={Neuron},
  volume={95},
  number={2},
  pages={245--258},
  year={2017},
  publisher={Elsevier}
}

@mastersthesis{czech2019deep,
	       title = { Deep Reinforcement Learning for Crazyhouse },
	       author = { Johannes Czech },
               year = { 2019 },
               type = { M.Sc. },
	       url = {https://github.com/QueensGambit/CrazyAra},
	       school = { TU Darmstadt },
	       pages = { 54 },
	       month = { 12 }
	       }

@article{rosin2011multi,
  title={Multi-armed bandits with episode context},
  author={Rosin, Christopher D},
  journal={Annals of Mathematics and Artificial Intelligence},
  volume={61},
  number={3},
  pages={203--230},
  year={2011},
  publisher={Springer}
}

@inproceedings{
Zhang2020Why,
title={Why Gradient Clipping Accelerates Training: A Theoretical Justification for Adaptivity},
author={Jingzhao Zhang and Tianxing He and Suvrit Sra and Ali Jadbabaie},
booktitle={International Conference on Learning Representations},
year={2020},
url={https://openreview.net/forum?id=BJgnXpVYwS}
}

@article{watkins1992q,
  title={Q-learning},
  author={Watkins, Christopher JCH and Dayan, Peter},
  journal={Machine learning},
  volume={8},
  number={3-4},
  pages={279--292},
  year={1992},
  publisher={Springer}
}

@misc{yu_ue_2018,
	title = {{NNUE} {Efficiently} {Updatable} {Neural}-{Network} based {Evaluation} {Functions} for {Computer} {Shogi}},
	url = {https://raw.githubusercontent.com/ynasu87/nnue/master/docs/nnue.pdf},
	publisher = {Ziosoft Computer Shogi Club},
	author = {Yu, Nasu},
	month = apr,
	year = {2018}
}

@article{gelly2011monte,
  title={Monte-Carlo tree search and rapid action value estimation in computer Go},
  author={Gelly, Sylvain and Silver, David},
  journal={Artificial Intelligence},
  volume={175},
  number={11},
  pages={1856--1875},
  year={2011},
  publisher={Elsevier}
}

@article{auer2002finite,
  title={Finite-time analysis of the multiarmed bandit problem},
  author={Auer, Peter and Cesa-Bianchi, Nicolo and Fischer, Paul},
  journal={Machine learning},
  volume={47},
  number={2-3},
  pages={235--256},
  year={2002},
  publisher={Springer}
}

@article{srivastava2014dropout,
  title={Dropout: a simple way to prevent neural networks from overfitting},
  author={Srivastava, Nitish and Hinton, Geoffrey and Krizhevsky, Alex and Sutskever, Ilya and Salakhutdinov, Ruslan},
  journal={The journal of machine learning research},
  volume={15},
  number={1},
  pages={1929--1958},
  year={2014},
  publisher={JMLR. org}
}

@misc{Havard2019,
  author = {Havard, Pierre},
  title = {The free KingBase Lite 2019 database},
  year = {2019},
  publisher = {kingbase-chess.net},
  howpublished = {\url{https://archive.org/details/KingBaseLite2019}},
  note = {Accessed: 2020-12-19}
}



%





\newpage
\section{Appendix}
\label{sec:appendix}
\begin{table}[h!]
\caption{Overview of the hyperparamter configuration used for our evaluation and to which configuration they belong to. The main set of hyperparmaters of PUCT search is shared among all other evaluated search modifications.}
\label{tab:hyperparams}
\begin{tabular}{lcl}
\toprule
\textbf{Hyperparameter} & \textbf{Value} & \textbf{Description} \\
$Q_{\epsilon}$ & $0.01$ & MCGS  \\
$Q_{\text{weight}}$ & $2.0$ & Q-value for move \\
$\epsilon_\text{greedy}$ & $0.01$ & Epsilon Greedy \\
$\epsilon_\text{checks}$ & $0.01$ & Enhance Checks\\
\hline
Threads & 2 & AlphaZero* \\
$c_{\text{puct-init}}$ & 2.5 & AlphaZero* \\
$c_{\text{puct-base}}$  & 19652 & AlphaZero* \\
$\epsilon_\text{Dir}$ & 0.0 & AlphaZero* \\
$\text{Node}_\tau$ & 1.7 & AlphaZero* \\
$\tau$ & $0.0$ & AlphaZero* \\
Mini-Batch-Size & 16 & AlphaZero* \\
Virtual Loss & 1.0 & AlphaZero* \\
Q-Value initialization & -1.0 & AlphaZero* \\
Q-Value range & $[-1, +1]$ & AlphaZero* \\
\bottomrule
\end{tabular}
\end{table}
\hfill
\fontsize{9}{11}
\begin{table}[h!]
\caption{Hardware and libraries used for our experiments.}
\resizebox{\columnwidth}{!}{%
\label{tab:hardware}
\begin{tabular}{ll}
\toprule
\textbf{Hardware / Software} & \textbf{Description} \\
GPU & NVIDIA GeForce RTX2070 OC\\
Backend & TensorRT-7.0.0.11, float16 precision\\
GPU-Driver & CUDA 10.2, cuDNN 7.6.5 \\
CPU & AMD Ryzen 7, 1700 8-coreprocessor×16\\
Operating System & Ubuntu 18.04.4 LTS \\
Tournament Environment & Cutechess 1.1.0 \\
\textit{CrazyAra} & \footnotesize{{500da21e0bd9152657adbbc6118f3ebbc660e449}} \\
\bottomrule
\end{tabular}}
\end{table}
\normalsize

\end{document}